\documentclass[journal]{IEEEtran}

\usepackage{hyperref} 
\usepackage{graphicx}
\DeclareGraphicsExtensions{.eps,.jpg,.pdf,.png,.ps}
\usepackage{subcaption}
\usepackage{cite}
\usepackage{amsmath,amssymb,mathtools}
\usepackage{siunitx} 

\begin{document}

\title{Computer vision-based framework for extracting geological lineaments from optical remote sensing data}

\author{Ehsan Farahbakhsh,
        Rohitash Chandra,
        Hugo K. H. Olierook,
        Richard Scalzo,
        Chris Clark,
        Steven M. Reddy,
        and R. Dietmar M\"{u}ller
\thanks{E. Farahbakhsh, R. Chandra and R. D. M\"{u}ller are with the School
of Geosciences, University of Sydney, Sydney, NSW 2006, Australia
(e-mail: e.farahbakhsh@sydney.edu.au, rohitash.chandra@sydney.edu.au, dietmar.muller@sydney.edu.au).}
\thanks{R. Scalzo is with the Centre for Translational Data Science, University of Sydney, Sydney, NSW 2006, Australia (e-mail: richard.scalzo@sydney.edu.au).}
\thanks{H. K. H. Olierook, C. Clark and S. M. Reddy are with the School of Earth and Planetary Sciences, Curtin University, Perth, WA 6845, Australia (e-mail: hugo.olierook@curtin.edu.au, c.clark@curtin.edu.au, s.reddy@curtin.edu.au).}}

\maketitle

\begin{abstract}
The extraction of geological lineaments from digital satellite data is a fundamental application in remote sensing. The location of geological lineaments such as faults and dykes are of interest for a range of applications, particularly because of their association with hydrothermal mineralization. Although a wide range of applications have utilized computer vision techniques, a standard workflow for application of these techniques to mineral exploration is lacking. We present a framework for extracting geological lineaments using computer vision techniques which is a combination of edge detection and line extraction algorithms for extracting geological lineaments using optical remote sensing data. It features ancillary computer vision techniques for reducing data dimensionality, removing noise and enhancing the expression of lineaments. We test the proposed framework on Landsat 8 data of a mineral-rich portion of the Gascoyne Province in Western Australia using different dimension reduction techniques and convolutional filters. To validate the results, the extracted lineaments are compared to our manual photointerpretation and geologically mapped structures by the Geological Survey of Western Australia (GSWA). The results show that the best correlation between our extracted geological lineaments and the GSWA geological lineament map is achieved by applying a minimum noise fraction transformation and a Laplacian filter. Application of a directional filter instead shows a stronger correlation with the output of our manual photointerpretation and known sites of hydrothermal mineralization. Hence, our framework using either filter can be used for mineral prospectivity mapping in other regions where faults are exposed and observable in optical remote sensing data.
\end{abstract}

\begin{IEEEkeywords}
Computer vision, Geological lineament, Landsat 8, Gascoyne Province.
\end{IEEEkeywords}

\section{Introduction}
\IEEEPARstart{D}{igital} satellite data with different spatial and spectral resolution are available for almost every locality on the Earth's land surface \cite{ElJanati2014,Thenkabail2015,VanderWerff2016,Hewson2017,DosReisSalles2017}. This enables the procurement of detailed information from surficial features and processes at different scales. Linear features are considered as one of the most important surficial features in different fields of study \cite{Hao2007,He2008,Pirasteh2013}. A linear feature is a two-dimensional, straight or slightly curved line, linear pattern or alignment of discontinuous patterns evident in an image, photo or map \cite{Wang1993}. Linear features represent the expression of some degree of linearity of a single or diverse grouping of both natural and cultural features \cite{Wang1993,Simonett1983}. Their specific physiographic characteristics make it possible to detect them due to the tonal change in digital satellite data \cite{Hashim2013}. Natural features include any linear features formed by natural processes such as geological lineaments, drainage networks, and vegetation alignments. Cultural features comprise man-made features such as road networks, railroads, and power line corridors \cite{Wang1993}.\\
The identification of linear features in remote sensing imagery can be complex since their spatial and spectral characteristics vary along their extent \cite{Wang1993}. A variety of techniques have emerged for extracting linear features from digital satellite data for different applications such as detecting road networks \cite{Valero2010,Singh2013}, stream networks \cite{Martinez2009,Paiva2015}, and geological lineaments \cite{Hashim2013,Marghany2010}. Geological lineaments are an expression of the underlying geological structure and include faults, dykes, shear zones and folds. Linear to curvilinear faults, dykes and shear zones are of particular interest in assisting mineral prospecting because of their association with hydrothermal mineralization \cite{BeiranvandPour2015,BeiranvandPour2016,Manuel2017} but their applications also extend to hydrogeological \cite{Bhuiyan2015,Dasho2017,Akinluyi2018,Takorabt2018} and tectonic studies \cite{Arian2015,Daryani2015,Masoud2017}. Faults, dykes and shear zones may be used to delineate major structural units, analyze of structural deformation patterns, and identify geological boundaries and uncover mineral deposits \cite{Glasser2009,Saadi2009,Raharimahefa2009,Ramli2010,BeiranvandPour2014}. Mineral deposits are commonly clustered around specific positions along deep crustal structures, which often have surface expressions in the absence of sedimentary or regolith cover \cite{Lund2011,Hein2013}. Understanding the structural mechanisms for hydrothermal ore deposits along deep crustal discontinuities are pivotal in terms of economic considerations but are still in dispute. Some researchers posit that fault bends are the primary control for ore clusters, while others believe that fault intersections are most important \cite{Lu2016}. Although, both cases are sometimes important in formation of hydrothermal ore deposits. Irrespective of the mechanism, detecting geological lineaments can help link surface lineament expressions to deep-seated structural discontinuities, which ultimately aids in mineral prospectivity mapping.\\
Although manual interpretation is effective at identifying geological lineaments \cite{Saadi2008,Saad2011}, computer vision techniques are required to make this process efficient. Computer vision is an interdisciplinary field that employs a wide range of algorithms for gaining a high-level understanding from digital images or videos. In other words, it seeks to automate tasks akin to what the human visual system can accomplish \cite{Ballard1982,Sonka2014}. Computer vision is concerned with automatic extraction, analysis and understanding of useful information from a single image or a sequence of images. It involves the development of a theoretical and algorithmic basis to achieve automatic visual understanding \cite{Szeliski2010}. Some of the applications of computer vision include systems for automatic inspection, detecting events, modelling objects and organizing information \cite{Szeliski2010}. In the geosciences, computer vision has diverse applications including surface modelling \cite{James2012}, rock type classification \cite{Patel2016}, motion analysis \cite{Gutierrez2002}, edge detection and extracting linear features from digital imagery \cite{Cross1988}.\\
The use of computer vision and satellite images to map geological lineaments has been particularly useful in regional scale studies that are inaccessible, unsafe or costly to navigate \cite{Karnieli1996,Shahzad2011}. Traditionally, lineament mapping is based on a visual or manual photointerpretation. Manual digitizing of lineaments is subjective, time consuming and expensive \cite{Rahnama2014a}. Therefore, a reliable and standard framework consisting of available computer vision techniques is needed. This framework must be able to work with different satellite images representing different spatial and spectral resolutions. The success of geological lineament extraction procedures depends on the sequence of applied techniques, and the reliability and accuracy of the edge detection and line extraction mechanisms \cite{Karantzalos2006}. Selection of computer vision techniques, appropriate edge detection methods and line extraction methods are pivotal in tectonic linear feature extraction because they exert significant influence on the accuracy of the final results.\\
In this study, we present a framework for detecting geological lineaments using computer vision techniques that include edge detection and line extraction methods. To demonstrate the validity of this framework, we use Landsat 8 satellite data and Shuttle Radar Topography Mission (SRTM) digital elevation models for extracting geological lineaments of the Yinnetharra 1:100,000 map sheet located in the Capricorn Orogen, Western Australia. Satellite data are subjected to dimensionality reduction, noise removal and lineament enhancement prior to edge detection and line extraction algorithms. The extracted geological lineaments are compared to our manual photointerpretation and geological mapping by the Geological Survey of Western Australia \cite{Johnson2012}. We also investigate the correlation between extracted geological lineaments and known mineral occurrences in the study area. The final product of our workflow is an efficient evidential data layer for detecting hydrothermal mineral deposits.\\
The rest of this paper is organized as follows. In section 2, geological setting of the study area is investigated. Section 3 reviews different computer vision techniques which have been applied in this study. Then, materials and the proposed framework for extracting geological lineaments are described in section 4. Results of applying the proposed framework are reported in section 5. The evaluation and limitations of the proposed framework are discussed in section 6. Finally, Section 7 concludes the paper with a summary of the future works.

\section{Geological setting of the study area}
\begin{figure*}
  \centering
  \includegraphics[width=\linewidth]{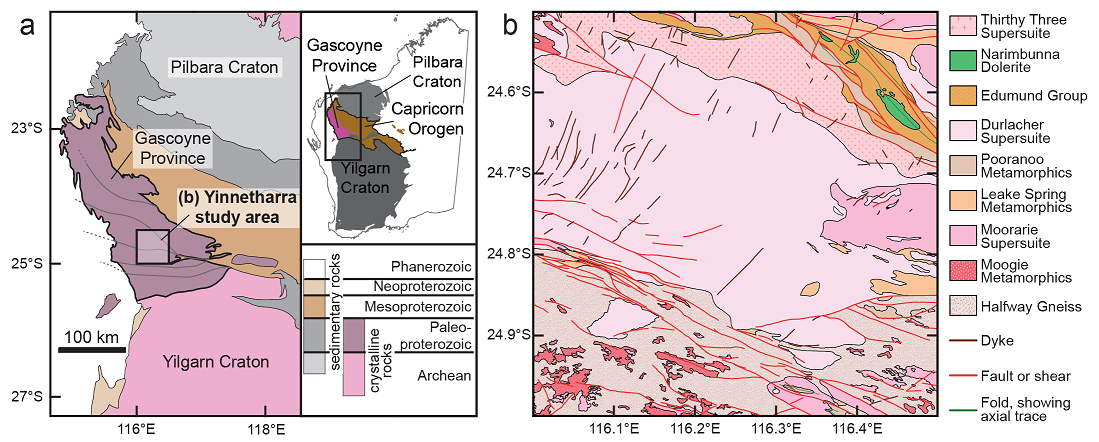}
  \caption{a) Map of the Capricorn Orogen showing the tectonic units and study area; b) simplified geological map of Yinnetharra 1:100,000 map sheet.}
  \label{figure01}
\end{figure*}

The Yinnetharra 1:100,000 map sheet is situated within the western portion of the Capricorn Orogen, known as the Gascoyne Province (Fig. \ref{figure01}) \cite{Johnson2013}. The Gascoyne Province, and the wider Capricorn Orogen, record the protracted amalgamation of the West Australian Craton and subsequent intracontinental tectono-thermal activity. Two main events have been found to contribute to forming the West Australian Craton. First, the ca. 2195--2145 Ma Ophthalmia Orogeny sutured the Glenburgh Terrane (comprised of the Halfway Gneiss) to the Pilbara Craton \cite{Rasmussen2005,Krapez2017}. The Opthalmia Orogeny was associated with the deposition of the Moogie Metamorphics, which was deposited into a foreland basin that formed as a response to the Glenburgh--Pilbara collision \cite{Johnson2013}. Second, the ca. 2005--1950 Ma Glenburgh Orogeny amalgamated the combined Pilbara Craton--Glenburgh Terrane with the Yilgarn Craton to form the West Australian Craton \cite{Johnson2013,Olierook2018}. The Glenburgh Orogeny was associated with two major Andean-type granitoid formations, the Dalgaringa and Bertibubba Supersuites, and several subduction-related basins \cite{Olierook2018,Johnson2011}. After unification, the Capricorn Orogen experienced at least five intracontinental tectono-magmatic events, each decreasing in severity of tectonic character and magmatism \cite{Johnson2017}. The first two events, the 1830--1780 Ma Capricorn Orogeny and 1680--1620 Ma Mangaroon Orogeny, were both associated with significant granitoid magmatism of the Moorarie and Durlacher Supersuites, respectively \cite{Sheppard2005,Sheppard2010}. Later events were predominantly amagmatic but were still associated with up to amphibolite-facies metamorphism and hydrothermal activity \cite{Sheppard2007,Korhonen2017}.\\
Both suturing and intracontinental tectonic events have developed a pervasive east--west striking structural fabric in the Gascoyne Province that has compartmentalized the region into several geological \lq{zones}\rq{} that share tectonic characteristics \cite{Sheppard2010}. In the south, zone and formation boundaries trend SW--NE whereas major structures are oriented NW--SE in the north, yielding a wedge-shaped geometry for the western margin of the Capricorn Orogen (Fig. \ref{figure01}). Compared to the rest of the Capricorn Orogen, the Gascoyne Province is relatively well exposed, but there are still significant regions of recent cover that hamper mineral exploration. Known precious and base metal mineralization predominantly occur along and are associated with structural discontinuities \cite{Johnson2013}. A crucial aspect to the mineralization potential in the Capricorn Orogen is the repeated reactivation of major structures over a billion years, which allows mineralization to be upgraded to economic scales \cite{Pirajno2004,Zi2015}.

\section{Computer vision techniques}
\subsection{Dimension reduction}
Dimension reduction is a machine learning method that reduces a set of random variables by obtaining a set of principle variables \cite{Zhang2010,Lafon2006}. Dimension reduction techniques can be divided into transformation methods (feature extraction) and feature selection methods \cite{Huang2007}. Transformation methods, such as Principal Component Analysis (PCA), Independent Component Analysis (ICA) and Minimum Noise Fraction (MNF), can separate significant and insignificant parts of the data in addition to reducing data dimensionality. Dimension reduction techniques are initially used to compress available information stored in multiple bands into a few bands. In other words, a multispectral image is converted into a greyscale image preserving maximum information. This technique can also be applied for different purposes such as statistical analysis and classification \cite{Lafon2006}.

\subsubsection{Principal component analysis}
Principal component analysis is an exploratory analysis method which aims to find a set of linearly uncorrelated components (principal components) which could serve as projections from the original data. The first principal component has the largest possible variance, and each succeeding component in turn has the next highest variance possible under the constraint that it must be orthogonal to the preceding components. The mathematical model of the PCA can be expressed as \cite{Yang2015}:

\begin{equation}
    \label{equation01}
    z = Px
\end{equation}

where $x$ is a vector consisting of $n$ rows of random variables $x_{1}$, $x_{2}$,..., $x_{n}$, each $x_{i}$ giving a particular datum that has zero mean, $P$ is the $n \times n$ standardized orthogonal transformation matrix or projection square matrix in which each projection vector (row vector) is constrained to be a unit vector, $z$ is a vector consisting of $n$ rows of random variables $z_{1}$, $z_{2}$,..., $z_{n}$ projected by $P$ from $x$.\\
Principal component analysis is one of the techniques that is applied to transform remote sensing data in order to reveal the most important features of an image \cite{Singh1985,Mackiewicz1993}. It is a technique for extracting a smaller set of variables with less redundancy from high-dimensional data sets in order to retain as much of the information as possible \cite{Eklundh1993,Liu2009}. Principal component analysis allows the determination of linear combinations of variables, feature extraction, dimension reduction, multidimensional data visualization and underlying variable identification \cite{Richards2013}. The principal components are a projection of the data onto principal axes, known as eigenvectors. A small number of principal components is often sufficient to account for the majority of the variance in the data \cite{Hall1997,Tobin2007}.

\subsubsection{Independent component analysis}
An alternative method for source separation is independent component analysis \cite{Comon1994,Hyvarinen1999,Ghahramani2004}. Unlike PCA, this method both decorrelates input signals and also reduces higher-order statistical dependencies \cite{Lee1998,Hyvarinen1999a}. Independent component analysis is a well-established statistical signal/data processing technique that aims at decomposing a set of multivariate signals into a base of statistically independent data-vectors with minimal loss of information \cite{Fiori2003,Ji2014}. Independent component analysis reveals hidden factors that underlie sets of random variables, measurements, or signals and it attempts to make the separated signals as independent as possible.\\
Independent component analysis has a wide range of applications in the field of signal processing, although it has been less considered as a common technique in image processing \cite{Benlin2008}. It is thought that lack of both comprehensive understanding of the ICA principle, and proper procedures to interpret the results generated by the ICA, can be one of the main reasons that ICA is under used in the geosciences, especially for multi- or hyper-spectral image processing. ICA is a special case of blind source separation that aims to separate source signals from mixture signals with or without little prior information about the source signals or the mixing process \cite{Cardoso1998}. The mixture model can be given by \cite{Yang2015}:

\begin{equation}
    \label{equation02}
    x = As
\end{equation}

where $x$ is a vector consisting of received mixture signals, $s$ is a vector that contains unknown source signals, and $A$ is an unknown full rank and invertible mixing matrix. In blind source separation models, the possible distribution of each source is unknown and we are not interested in, and do not have prior information about, any source. However, the lack of prior knowledge about the mixture is compensated for by a statistically strong and often physically plausible assumption of statistical independence between the source signals \cite{Cardoso1998}. This assumption provides a possible solution: finding some signals that are independent from each other which can be mixed to reproduce the observed signals. Therefore, the aim of ICA is to find a set of uncorrelated components as independent as possible from each other. Accordingly, the decomposition model can be given as \cite{Yang2015}:

\begin{equation}
    \label{equation03}
    s = Wx = WAs
\end{equation}

where $W$ is the unknown unmixing square matrix to be determined. For simplification of the model without loss of generality, the independent components and the mixture signals are always assumed to have zero mean and unit variance. This assumption yields that there is no variance ranking of the independent components. Via various estimators of independence, there are many mature algorithms available for implementing ICA. In this study, the Fast ICA, a fixed-point algorithm that uses an approximation of negentropy as measurement of independence, is chosen for data processing due to its computing efficiency, flexible parameters and robustness \cite{Hyvarinen1999a}.

\subsubsection{Minimum noise fraction}
Minimum noise fraction is similar to the PCA transform and is an effective technique for reducing a large multi-dimensional data set into a fewer number of components that contain the majority of information \cite{Green1988}. Unlike a PCA transform, the resulting axes or components are not necessarily orthogonal, but are ordered by decreasing signal-to-noise ratio. Moreover, MNF is applied to isolate noise from signal in a dataset and to determine inherent dimensionality of an image. Minimum noise fraction is also able to reduce computational requirements for subsequent processing \cite{Boardman1994}.\\
The MNF transform implemented in this study involves two cascaded PCA transformations. The first transformation, based on an estimated noise covariance matrix, decorrelates and rescales the noise in the data. The second step is a standard PCA transformation of the noise-reduced data. There can be as many components as there are input bands; however, each component, starting from the first, describes less and less of the overall variance of the data set. Typically, only a small number of components are required to describe most of the information for the entire data set. The contribution of each component to the overall information in a multivariate data set (i.e., multispectral or hyperspectral imagery) is measured by an eigenvalue. A larger eigenvalue indicates that a component contains more information from the data set. The contribution of each band to each component is measured by an eigenvector, which can be interpreted akin to a correlation coefficient. Unlike a PCA transform, the resulting axes (components) are not necessarily orthogonal but are ordered by decreasing signal-to-noise ratio \cite{Harris2005}.

\subsection{Image enhancement}
The image enhancement techniques are applied to increase the quality of image interpretation based on enhancing edges of features prior to their detection and extraction. Different types of filtering such as adaptive and convolutional filtering are part of this suite of techniques.\\
Adaptive filtering uses the standard deviation of those pixels within a local box surrounding each pixel to calculate a new pixel value. Typically, the original pixel value is replaced with a new value calculated based on the surrounding valid pixels which satisfy the standard deviation criteria \cite{Crow2008}. The adaptive filters are particularly adept at preserving image sharpness and detail while suppressing noise. Lee filter is an adaptive filter which is used to smooth noisy and speckled data that have an intensity related to the image scene and that also have an additive and/or multiplicative component \cite{Lee1980}.\\
Convolutional filters yield images in which brightness value at a given pixel is a function of weighted average of surrounding pixels' brightness. Convolution of a user-selected kernel with the image array returns a new, spatially filtered image. Different kernel size and values produce different types of filters. Standard convolutional filters include high pass, low pass, Laplacian, directional, median, Sobel and Roberts \cite{Richards2013,Haralick1987}. In the proposed framework for extracting geological lineaments, we have applied median and compared directional and Laplacian filters. A median filter is applied for smoothing the image, while preserving edges larger than the kernel dimensions and also to remove speckling noise \cite{Chan2005}. This low frequency filter replaces each center pixel with the median value within the neighborhood specified by the filter size \cite{Haralick1987}. A directional filter is a first derivative edge enhancement filter that selectively enhances image features having specific direction components (gradients) \cite{Zhang2006a}. The sum of the directional filter kernel elements is zero. The result is that areas with uniform pixel values are zeroed in the output image, while those that are variable are presented as bright edges. A Laplacian filter is a second derivative edge enhancement filter that operates without regard to edge direction \cite{Lee1990}. Laplacian filtering emphasizes maximum values within the image using a kernel with a high central value typically surrounded by negative weights in north--south and east--west directions and zero values at the kernel corners \cite{Haralick1987}.

\subsection{Lineament extraction}
Linear features can be divided into two major subclasses that include edges and lines on a digital greyscale image. An edge in an image is an abrupt discontinuity in image brightness, which may result from surface boundaries, shadow boundaries or changes in surface reflectance. A line in an image is a digital valley or ridge of image brightness \cite{Wang1993}. Lines may be formed, for example, by roads, rivers, and vegetation alignments, as well as by geological faults and joints. Lineament extraction methods can be conducted via manual photointerpretation by an expert or (semi-)automatic detection using computer vision techniques \cite{Vassilas2002}. The automatic methods have resulted in a more efficient lineament extraction process \cite{Tripathi2000,Masoud2006,Masoud2011}. A lineament extraction process comprises two main steps, namely edge detection and line extraction.

\subsubsection{Edge detection}
In general, automatic lineament extraction methods are based on edge detection techniques that enhance the pixels at the edges on an image, instead of directly extracting edge contours. An edge in an image is defined as a boundary or contour at which a significant change occurs in some physical aspect of the image. Linear operators can detect edges through the use of masks that represent the ideal edge steps in various directions \cite{Ali2001}. They can detect lines and curves in much the same way. Traditional edge detectors were based on a rather small neighborhood, which only examined each pixel's nearest neighbor \cite{Cyganek2003,Awad2008}. This may work well but due to the size of the neighborhood that is being examined, there are limitations to the accuracy of the final edge. These local neighborhoods will only detect local discontinuities, and it is possible that this may cause false edges to be extracted. A more powerful approach is to use a set of first or second difference operators based on neighborhoods having a range of sizes and combine their outputs, so that discontinuities can be detected at various scales \cite{Ballard1982}.\\
Laplace, zero-crossing and gradient operators are the primary edge detection techniques \cite{Ali2001,Patel2011}. Laplace operators compute some quantity related to the divergence of the intensity surface gradient of the greyscale images. Zero-crossing operators determine whether or not the digital Laplacian or the estimated second direction derivative has a zero-crossing within the pixel \cite{Ali2001}. Gradient operators compute some quantity related to the magnitude of the slope of the greyscale image wherein the image pixel values are noisy discretized samples. The major drawback of such an operator is the fact that determining the actual location of the edge and slope turnover point is difficult \cite{Ali2001}. The frequency and connectivity of the extracted edges by common edge detection methods are strongly affected by the type and spatial resolution of the source datasets, signal-to-noise ratio and the parameters of the edge detection methods \cite{Zhang2006}. In other words, results of most edge detection methods involve fragmented edges and should be ultimately interpreted visually.\\
An alternative technique is known as Canny edge detection, which is based on computing the squared gradient magnitude \cite{Canny1986}. Local maxima of the gradient magnitude that are above some threshold are identified as the edges. The advantages of the Canny edge operator are deriving an optimal operator in the sense that minimizes the probability of multiply detecting an edge, minimizing the probability of failing to detect an edge, and minimizing the distance of the reported edge from the true edge. There is a tradeoff between detection and localization; the more accurate the detector the less accurate the localization and vice-versa \cite{Ali2001}. An objective function has been designed to achieve some optimization constraints including maximizing the signal-to-noise ratio to give perfect detection which favors the marking of true positives; achieving perfect localization to accurately mark the edges; minimizing the number of responses to a single edge which favors the identification of true negatives and prevents marking non-edges \cite{Ali2001}.

\subsubsection{Line extraction}
After edge detection, additional processing must be performed in order to remove false edge responses and to link gaps between edges, ultimately turning isolated edges into lines. This processing turns the linearization of the edge pixels into continuous contours, using certain criteria such as closeness or some specific geometric property such as the degree of curvature \cite{Ghita2002}.\\
Edge-linking methods can be classified into two categories, namely local processing methods and global processing methods \cite{Rahnama2014}. In local processing methods such as the Hough transform and gradient-based methods, edge pixels are grouped to form edges by considering each pixel's relationship to any neighboring edge pixels \cite{Ghita2002}. This method is suitable to link edge pixels in situations where the shape of the edge is unknown. The Hough transform is a well-known technique which tolerates noise and discontinuities in an image but it has limitations such as high computing time, unwieldy memory requirements and limited capability in preserving edge pixel connectivity \cite{Hashim2013}. The global processing methods such as pixel connectivity-edge linking use all edge pixels. Pixels displaying similarities such as same edge geometry are used to find the best fit of a known shape. The global methods do not need to connect the edge pixels. However, these methods may miss small pieces of edges or noise pixels may wrongly be handled as edge segments \cite{Yang2011}. The pixel connectivity-edge linking algorithms have been widely used in applications which involve extraction of continuous line segments \cite{Hashim2013}.

\section{Materials and methods}
\subsection{Landsat 8 satellite data}
The Landsat 8 satellite carries two instruments including the Operational Land Imager (OLI) and Thermal Infrared Sensor (TIRS). The OLI comprises nine spectral bands with a spatial resolution of 30 meters for bands 1 to 7 and 9. The resolution for band 8 (panchromatic) is 15 meters. The Thermal Infrared Sensor (TIRS ) provides two thermal bands 10 and 11 that provide more accurate surface temperatures but are less useful for geological purposes \cite{Irons2012,Loveland2012}. These sensors both provide improved signal-to-noise radiometric performance quantized over a 12-bit dynamic range that enables better characterization of land cover state and condition compared to other Landsat satellites \cite{Reuter2010}. Moreover, geological features and the geomorphological framework are more readily distinguished.\\
The cloud-free Landsat 8 OLI level-2 data product (surface reflectance) is used in this study and was obtained from USGS EarthExplorer (\url{https://earthexplorer.usgs.gov}). The considered scene was acquired on 13 February 2018. Surface reflectance products provide an estimate of the surface spectral reflectance as it would be measured at ground level in the absence of atmospheric scattering or absorption \cite{Vermote2016}. The scene was pre-georeferenced to UTM zone 50 south projection with the WGS-84 datum, but all the outputs of this study are provided in GDA-94. Bands 1 and 9 are not used in this study, because they have been designed for retrieving atmospheric aerosol properties and detecting cirrus cloud, respectively \cite{Adiri2016}.

\subsection{Structural geological maps}
Structural geological maps represent the expression of underlying geological structures exposed at the Earth's surface \cite{Lisle2003}. Structural analysis involves the description of the structural geometry of a deformed field area, kinematic analysis and dynamic analysis \cite{Lageson2009}. Descriptive structural analysis is obtained through detailed field mapping, yielding positions, orientations and lengths of linear structures, including faults, dykes and fold axes. No geological map is ever complete but mapping by the Geological Survey of Western Australia (GSWA) at 1:100,000 is an appropriate benchmark for assessing the validity of our final workflow results.

\subsection{Manual photointerpretation}
Photointerpretation can be defined as the dialectic and interdisciplinary integration of personal experience, reasoning, expertise and ground truth \cite{Rokos1995}. In manual photointerpretation, fundamental photo-recognition elements of visual interpretation, as well as their appropriate combinations, are used \cite{Soille2002}. In remote sensing data, manual interpretation is restricted to analyzing only one or three channels of the satellite data at a time. Further, our visual acuity does not allow us to identify all spectral differences in imagery. A human interpreter can only detect and evaluate noticeable differences in the imagery and she/he cannot carry out repeatable interpretation work. The manual approach suffers from its inability to deal quickly with a large quantity of image data. Therefore, visual interpretation of digital imagery provided by remote sensing platforms does not allow full exploitation of the data provided \cite{Lillesand2014}. Nevertheless, manual photointerpretation is another technique that can be used to validate the proposed semi-automated workflow. Thus, we used manual photointerpretation to generate a geological lineament map that can be readily contrasted with the semi-automated extracted lineaments.\\
The simplest way and first step to work with multispectral data such as Landsat 8 is to display them as a False Color Composite (FCC). False color composite images have diverse applications in geological studies such as discriminating textural characteristics of igneous rocks from those of sedimentary rocks. Structural features are readily recognizable by applying a proper RGB color combination of the available bands. Several FCC images can be generated using the OLI bands, but the ones which assist to discriminate different lithological units, alteration zones and structural features are the most important in manual photointerpretation. Therefore, the optimum FCC image plays an important role in prospecting hydrothermal mineral deposits without using computer vision techniques \cite{Liu2013,Mwaniki2015}. There are different statistical methods for determining the best FCC image including optimum index factor, Sheffield index and correlation index \cite{Beauchemin2001}. In this study, the correlation index -- known as the most efficient index -- is applied for determining the best FCC image using the MATLAB code available at \url{https://github.com/intelligent-exploration/IP_MinEx}.

\subsection{Computer vision-based framework for semi-automated extraction of geological lineaments}
\begin{figure}
  \centering
  \includegraphics[width=\linewidth]{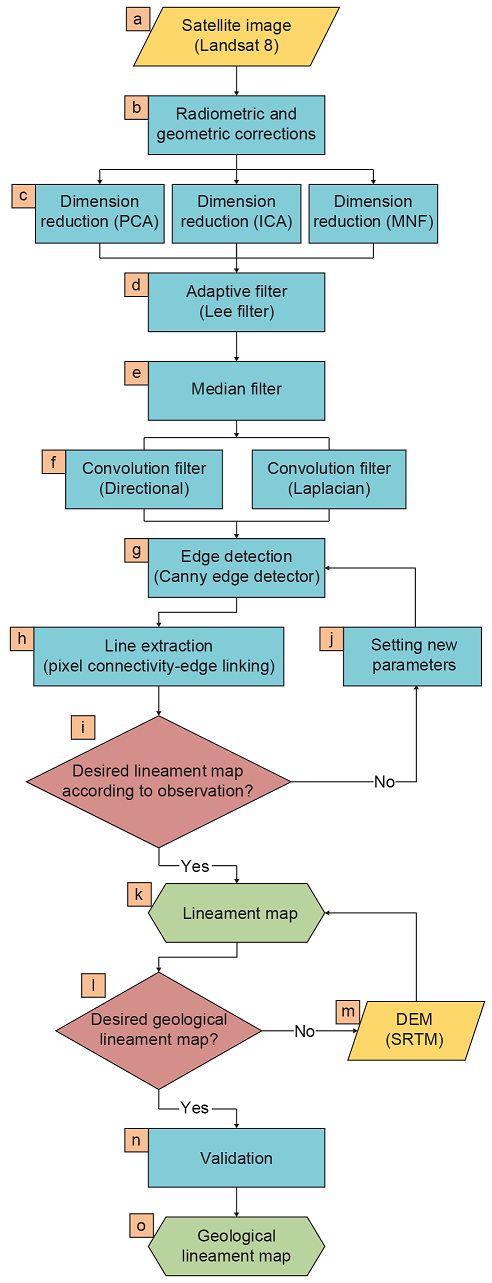}
  \caption{Methodology flowchart of this study for semi-automatic extraction of the geological lineaments. Input data to this flowchart include a digital satellite image (Landsat 8) and a digital elevation model (SRTM).}
  \label{figure02}
\end{figure}

A detailed computer vision-based framework for semi-automatic extraction of geological lineaments from optical remote sensing data is presented in Fig. \ref{figure02}. This framework starts with acquiring satellite images (Fig. \ref{figure02}a). According to the level of satellite images, radiometric and geometric corrections are applied (Fig. \ref{figure02}b). To transform satellite images from multiple bands into a greyscale (single-band) image, three dimension reduction techniques are used and compared (PCA, ICA and MNF) using ENVI 5.3 software (Fig. \ref{figure02}c). After image transformation, Lee and median filters (Fig. \ref{figure02}d, e) are applied respectively to the output component of the three dimension reduction techniques with the highest eigenvalue. To enhance edges, directional and Laplacian filters (Fig. \ref{figure02}f) are subsequently applied on greyscale images using ENVI 5.3. The directional filter is applied using a $3 \times 3$ kernel in four directions with azimuths of \ang{0}, \ang{45}, \ang{90} and \ang{135}. Using these azimuths, the directional filter visually enhances the edges striking N--S, NE--SW, E--W and NW--SE, respectively. The Laplacian filter uses a $3 \times 3$ kernel with a value of 4 for the center pixel and values of -1 for the N--S and E--W pixels.\\
Edge detection is performed using a Canny edge detector applied on the final greyscale image (Fig. \ref{figure02}g). Canny edge detection is preferred over Laplace, zero-crossing or gradient operators because it is resistant to a noisy environment and signals can be enhanced with respect to the noise ratio by a non-maxima suppression method which results in one pixel wide ridges as the output. Lines are then extracted from the binary image produced by the Canny edge detection process using pixel connectivity-edge linking (Fig. \ref{figure02}h). Pixel connectivity-edge linking is preferred over Hough transform or gradient based methods because all pixels identified as linear edge pixels are maintained \cite{Guru2004,Vucinic2010}.\\
The edge detection and lineament extraction process relies on setting six different thresholds by the user utilizing PCI Geomatica 2016 software, including filter radius, edge gradient, curve length, line fitting error, angular difference and linking distance (Fig. \ref{figure02}j; Table \ref{table01}). Filter radius and edge gradient threshold are effective for the edge detection and the other thresholds are applied for the line extraction. Several combinations of the thresholds within the proposed ranges in Table \ref{table01} were evaluated and various validation criteria as well as the ground truth were applied to reach the optimum values which gave a satisfactory result. The Canny edge detection and pixel connectivity-edge linking methods are applied separately on every four output images after implementing directional filter (Fig. \ref{figure05}a--\ref{figure05}d) for extracting the lineaments. The extracted lineaments from each image are combined and a comprehensive lineament map which covers major directions is generated.

\begin{table*}
  \centering
  \resizebox{\linewidth}{!}{%
  \begin{tabular}{|c|c|c|c|c|c|}
    \hline
    Threshold & Unit & Description & Proposed range & Proposed value (directional filter) & Proposed value (Laplacian filter) \\
    \hline
    Filter radius & Pixel & Filter radius used in the edge detection & 3-–8 & 5 & 5 \\
    \hline
    Edge gradient & - & Edge gradient value used in the edge detection & 10-–70 & 50 & 10 \\
    \hline
    Curve length & Pixel & Minimum length of a curve to be taken as a lineament & 10-–50 & 50 & 50 \\
    \hline
    Line fitting error & Pixel & Tolerance allowed in curve fitting to form a polyline & 2-–5 & 5 & 5 \\
    \hline
    Angular difference & Degree & Angle not to be exceeded between two polylines to be linked & 3-–20 & 10 & 10 \\
    \hline
    Linking distance & Pixel & Maximum distance between two polylines to be linked & 10–-50 & 50 & 50 \\
    \hline
  \end{tabular}}
  \caption{Thresholds applied for the edge detection and line extraction in this study \cite{Hashim2013,Adiri2017}.}
  \label{table01}
\end{table*}

The filter radius is initially used to determine the number of pixels to be considered around each pixel for applying the Canny edge detector. The edge gradient threshold then specifies the values of pixels to be taken as the edges and those remaining as background. Applying these two steps results in a binary map product. According to the curve length, this binary map undergoes a contour number reduction in order to leave just those referring to curves. In other words, a curve with a number of pixels smaller than that indicated by the threshold will not be taken into consideration. The resulting curves generate polylines if they fall within the tolerance defined by the line fitting error. Finally, two polylines in the simplest case will bind to form a lineament if their two end points form an angle respecting the value specified by the angular difference and the distance by the linking distance \cite{Adiri2017}.\\
According to the spatial resolution of the OLI imagery, streams are the most important cultural linear features which can be mistaken for geological linear features. Here, we used Shuttle Radar Topography Mission (SRTM) data with a spatial resolution of 30 m to delineate streams using the Esri Arc Hydro Tools (Fig. \ref{figure02}m). The map presented in Fig. \ref{figure03}b is applied to convert a lineament map (Fig. \ref{figure02}k) to a geological lineament map with minimal cultural lineaments (Fig. \ref{figure02}l--o). A buffer zone with a radius of 5 pixels (150 m) surrounding the streams is created to prevent errors caused by shifting pixels in SRTM digital elevation model. Lineaments that overlap the buffer zone by more than half their length are removed. Therefore, geological rather than cultural lineaments are more likely to be preserved in the final product (Fig. \ref{figure02}l--o).

\section{Results}
\subsection{Optimum false color composite image}
\begin{figure}
  \centering
  \includegraphics[width=\linewidth]{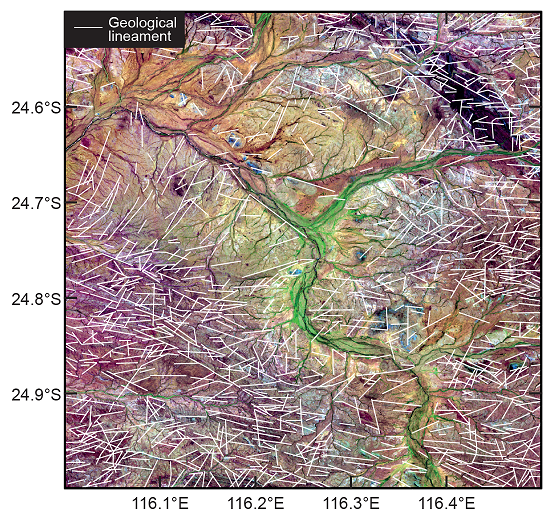}
  \caption{Extracted geological lineaments using manual photointerpretation draped over false color composite (RGB 752) image of the study area generated by the OLI bands. In this image, different colors imply the presence of different lithological units.}
  \label{figure03}
\end{figure}

The application of correlation index identifies RGB 752 as the best FCC image for visually discriminating lithological units and structural features, and manual photointerpretation of the study area (Fig. \ref{figure03}) \cite{Kamel2015,Abdelaziz2018}. Based on the ranking of available FCC images, RGB 753, 652 and 653 are alternatives. According to Fig. \ref{figure03} and extracted geological lineaments using manual photointerpretation, principal structural linear features strike NW--SE with subordinate lineaments striking NE--SW in the Yinnetharra region, akin to the dominant structural trend identified from geological mapping (Fig. \ref{figure01}).

\subsection{Comparison of dimension reduction transformations}
\begin{figure*}
  \centering
  \includegraphics[width=\linewidth]{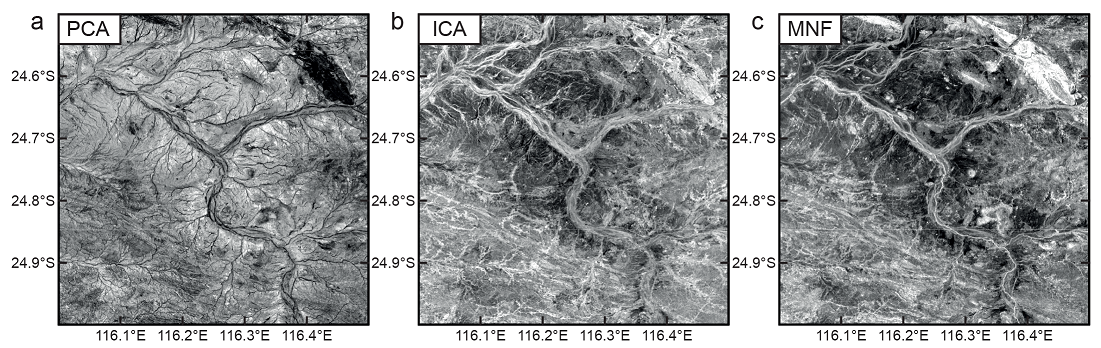}
  \caption{Output components of the a) PCA, b) ICA and c) MNF techniques with the highest eigenvalue. These components have been polished by the Lee and median filters respectively.}
  \label{figure04}
\end{figure*}

The output component of the PCA, ICA and MNF with the highest eigenvalue (including polishing using Lee and median filters on each, Fig. \ref{figure02}) yields greyscale images presented in Fig. \ref{figure04}. Geological lineaments are recognizable as dark and bright lines in the three output components (Fig. \ref{figure04}). Pixel values of the PCA and ICA components show negatively skewed distributions, while the MNF component shows a positively skewed distribution. The range of pixel values in the PCA component is much higher than the ICA and MNF components. Comparison of the components shows that dark and bight pixels as two anomalous populations of the pixel value distribution have been better separated in the MNF component.

\begin{figure*}
  \centering
  \includegraphics[width=\linewidth]{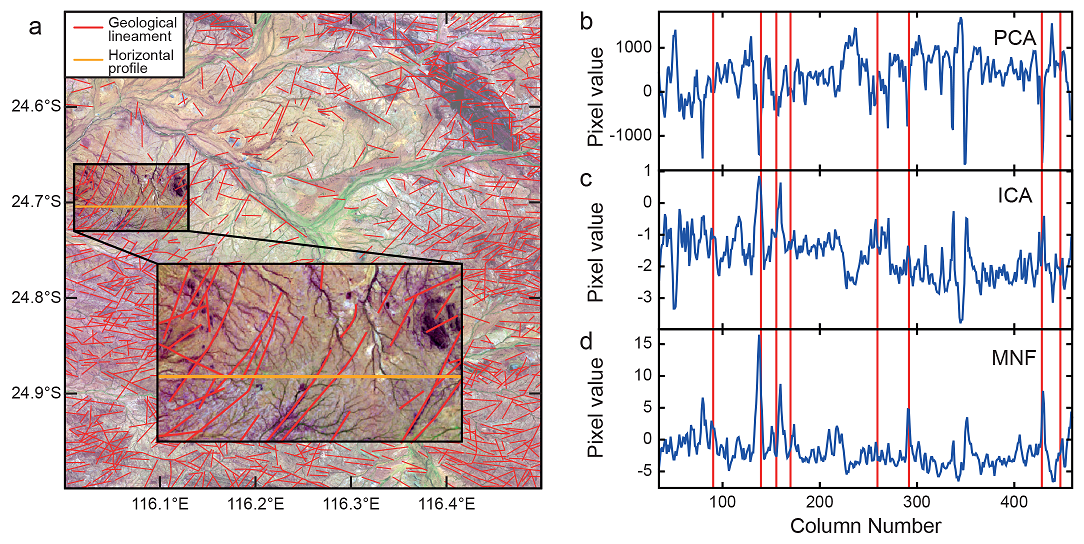}
  \caption{a) Geological lineament map provided by manual photointerpretation draped over a false Landsat 8 color composite image (RGB 752). The investigated horizontal profile in this map is shown in orange. b–d) The changes of pixel values relevant to the PCA, ICA and MNF components with the highest eigenvalue across the profile, respectively.}
  \label{figure05}
\end{figure*}

Comparison of the different dimension reduction techniques with manual photointerpretation shows that the MNF component is more robust than the PCA and ICA components for extracting geological lineaments (Fig. \ref{figure05}). Focusing on a horizontal profile across a small sector in the west of the Yinnetharra map sheet reveals that the number of peaks (i.e., positive anomalous pixels) on the profile obtained from the MNF component correlates well with the intersected geological lineaments identified via manual photointerpretation. The MNF profile shows a relatively uniform background signal with strongly pronounced positive anomalies. In contrast, the graphs obtained from the PCA and ICA components show intense fluctuation across the profile, showing a poorer correlation to the lineaments identified via manual photointerpretation.

\subsection{Comparison of directional and Laplacian filters}
\begin{figure*}
  \centering
  \includegraphics[width=\linewidth]{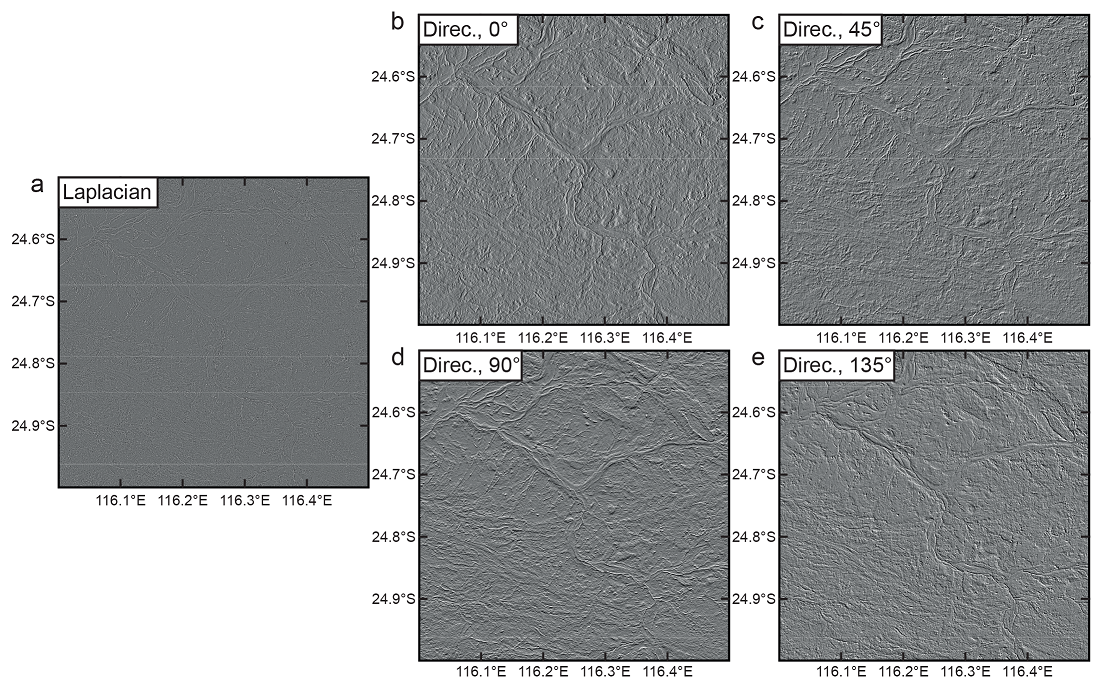}
  \caption{Images obtained by applying a) a Laplacian filter and a directional filter in four directions with azimuths of b) \ang{0}, c) \ang{45}, d) \ang{90}, e) \ang{135} on the output component of the MNF technique with the highest eigenvalue.}
  \label{figure06}
\end{figure*}

Application of directional and Laplacian filters on each of the outputs shown in Fig. \ref{figure04} reveals geological and cultural lineaments (Fig. \ref{figure06}). Streams and other large linear features can be readily identified in filtered images. However, there are some differences between the outputs of applying directional and Laplacian filters in terms of enhancing some of the geological lineaments. For example, the Laplacian filter failed to enhance most of the NW--SE striking geological lineaments located in the southwest of the study area, which are related to high number of hydrothermal mineral deposits. Applying a directional filter in different azimuths enhances geological structures striking in different directions and the results show that it is more robust in enhancing small lineaments compared to the Laplacian filter. Small NE--SW striking lineaments located in the northeast of the study area are probably mafic dykes (Mundine Well Dyke Swarm \cite{Wingate2000}), and these are better enhanced using the directional filter compared to the Laplacian filter (Fig. \ref{figure06}a).

\subsection{Extraction of geological lineaments}
\begin{figure}
  \centering
  \includegraphics[width=\linewidth]{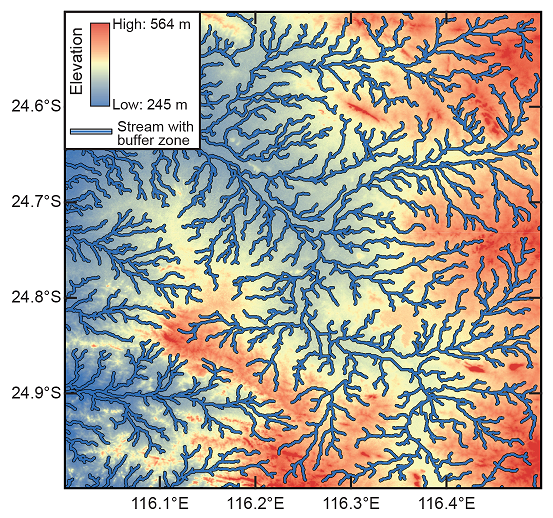}
  \caption{Extracted streams using SRTM digital elevation model.}
  \label{figure07}
\end{figure}

\begin{figure*}
  \centering
  \includegraphics[width=0.9\linewidth]{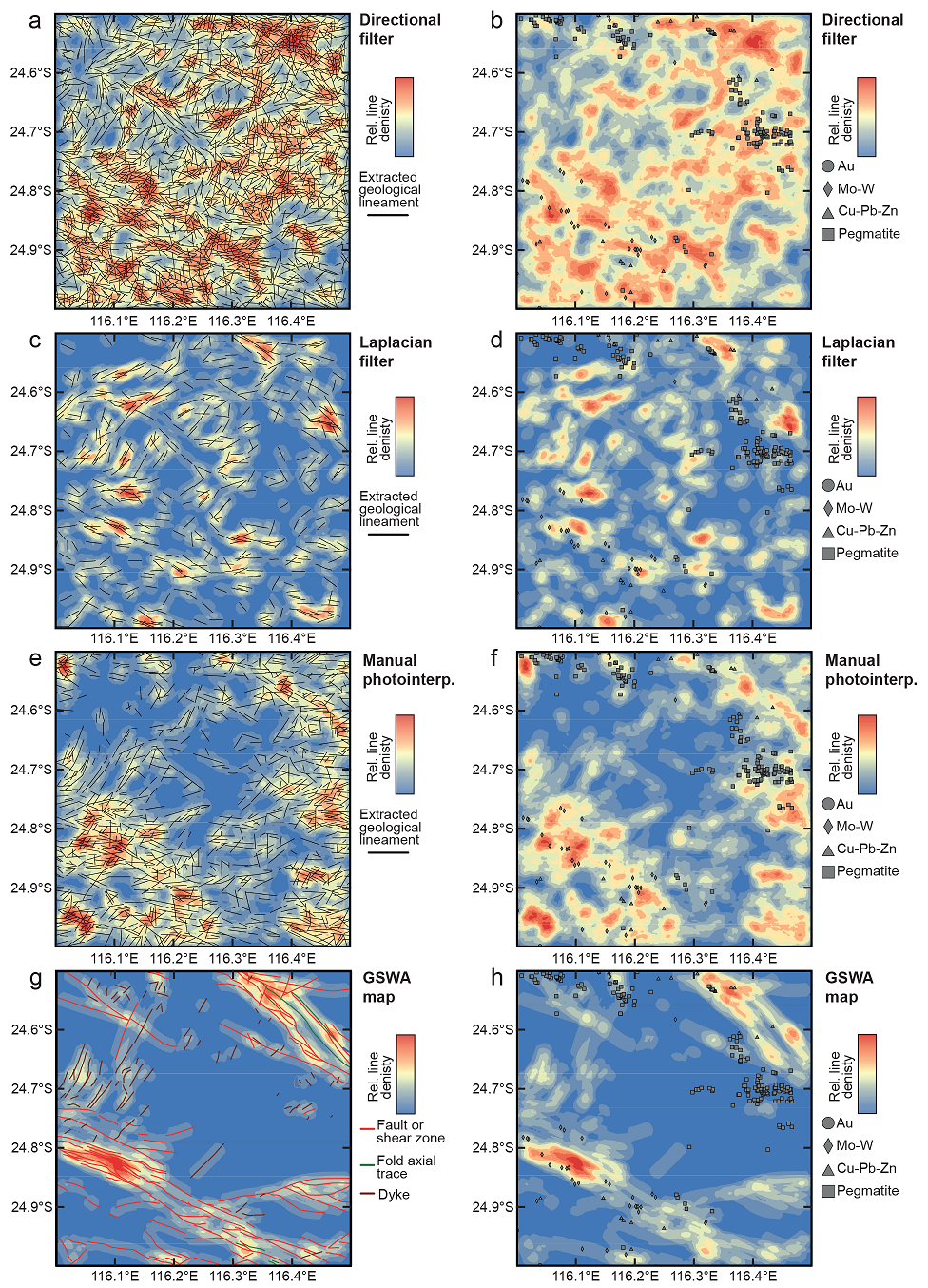}
  \caption{Superposition of geological lineaments and hydrothermal mineral occurrences on density maps resulted from the proposed framework using the MNF component improved by the a, b) directional and c, d) Laplacian filters; e, f) geological lineaments mapped by the manual photointerpretation and g, h) the GSWA. Density maps are created by summing the lineament length available in a defined grid size. For computational practicality, the size of each grid is set to 10 pixels (300 m) and the search radius is set to 50 pixels (1500 m). Lineament density values are taken into fuzzy space using a linear function, which yields values between 0 and 1 that makes it easier to compare them.}
  \label{figure08}
\end{figure*}

The most recognizable cultural linear features are streams, which may be mistaken for geological linear features (Fig. \ref{figure07}). Removal of streams and their buffer zones yield different geological lineament maps for directional and Laplacian filters (Fig. \ref{figure08}). To aid visualization and validation, a lineament density map is produced for each of the filters as well as the manual photointerpretation and GSWA lineament maps to analyze the dispersion pattern of the lineaments (Fig. \ref{figure08}).

The geological lineament map provided by the directional filter shows a significantly higher number of the lineaments compared to the map provided by the Laplacian filter (Fig. \ref{figure08}). Comparison of these with the result of manual photointerpretation and the GSWA map shows that the major trends of the geological lineaments have been displayed correctly. The map provided by the Laplacian filter is more correlated with the GSWA map but the density map of the geological lineaments extracted using the directional filter is highly correlated with hydrothermal mineral occurrences (Fig. \ref{figure08}). Areas with high density values located in the northeast and southwest of the study area (Fig. \ref{figure08}e) are associated with igneous granitic units, which can host diverse types of economic mineralization \cite{Candela1992}.

\subsection{Orientation analysis of extracted geological lineaments}
\begin{figure*}
  \centering
  \includegraphics[width=0.8\linewidth]{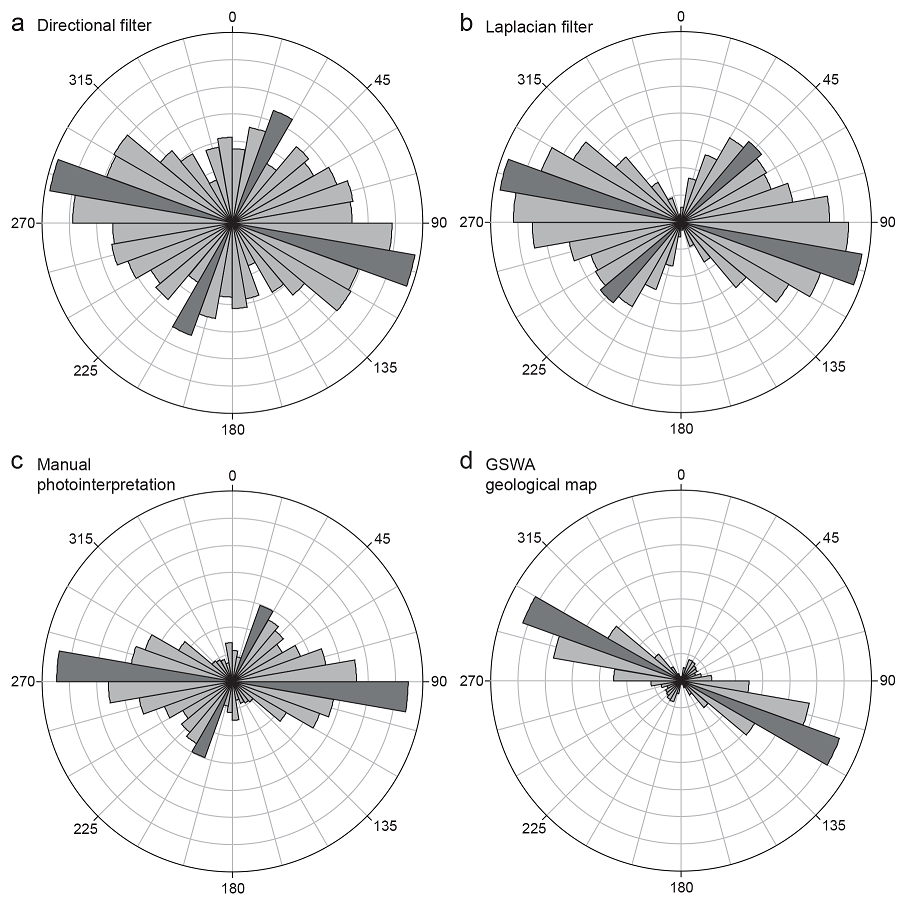}
  \caption{Rose diagrams showing the number and orientation of the geological lineaments extracted by the proposed framework using a) directional and b) Laplacian filters applied on the MNF component. As a means of validation, rose diagrams of the geological lineaments mapped by c) manual photointerpretation and d) the GSWA are also shown.}
  \label{figure09}
\end{figure*}

The geological lineaments extracted via directional and Laplacian filters (Fig. \ref{figure06}) reveal similar primary azimuth directions of \ang{100}--\ang{110}, which account for almost 10\% and 12\% of all extracted structures, respectively (Figs. \ref{figure08}, \ref{figure09}, Table \ref{table02}). The directional and Laplacian filters show subtle differences in subordinate populations, striking at \ang{20}--\ang{30} and \ang{40}--\ang{50}, respectively (Fig. \ref{figure09}, Table \ref{table02}).\\
The NW--SE striking lineaments are strongly correlated with the GSWA map and manual photointerpretation (Fig. \ref{figure09}). The geological lineaments extracted using manual photointerpretation shows a major and minor azimuth of \ang{90}--\ang{100} and \ang{20}--\ang{30}, accounting for almost 15\% and 7\% of all extracted geological lineaments, respectively (Fig. \ref{figure09}c, Table \ref{table02}). Geological mapping by the GSWA reveals a major azimuth of \ang{110}--\ang{120} that accounts for almost 26\% of all extracted geological lineaments (Fig. \ref{figure09}d, Table \ref{table02}). There is a far broader spread of NW--SE trending structures for the extracted lineaments using remote sensing data compared to those mapped by the GSWA. The subordinate NE--SW striking features identified via semi-automated lineament extraction are well represented in the manual photointerpretation but are comparatively rare in the GSWA map (Figs. \ref{figure08}, \ref{figure09}).

\begin{table*}
  \centering
  \begin{tabular}{|c|c|c|c|c|}
    \hline
    Method & Azimuth of the major strike & Percentage & Azimuth of the minor strike & Percentage \\
    \hline
    Proposed framework using directional filter & 100--110 & 10 & 20--30 & 6 \\
    \hline
    Proposed framework using Laplacian filter & 100--110 & 12 & 40--50 & 6 \\
    \hline
    Manual photointerpretation & 90--100 & 15 & 20--30 & 7 \\
    \hline
    GSWA geological lineament map & 110--120 & 26 & - & - \\
    \hline
  \end{tabular}
  \caption{Comparison of different geological lineament maps generated by applying different methods.}
  \label{table02}
\end{table*}

\subsection{Correlation of geological lineaments with hydrothermal mineral occurrences}
\begin{figure}
  \centering
  \includegraphics[width=\linewidth]{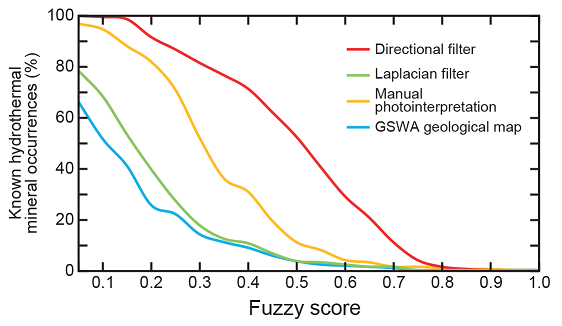}
  \caption{Spatial association between hydrothermal mineral occurrences and different geological lineament density maps. Lineament density values are taken into fuzzy space using a linear function. The \emph{x} axis shows the fuzzy thresholds from 0 to 1. The \emph{y} axis shows the percentage of known hydrothermal mineral occurrences that are placed in anomalous areas according to the thresholds shown in \emph{x} axis.}
  \label{figure10}
\end{figure}

A strong spatial association is observed between geological lineaments (e.g., faults, dykes) and hydrothermal mineral occurrences (Fig. \ref{figure08}). Converting geological lineaments into a density pattern (Fig. \ref{figure10}) further supports the notion that geological lineaments and hydrothermal mineralization are co-located. A density map of the extracted geological lineaments using a directional filter on the MNF component shows the highest correlation with hydrothermal occurrence locations (i.e., the highest area under curve in Fig. \ref{figure10}). The result of applying a directional filter shows an even better correlation with the hydrothermal mineralization compared to the result of manual photointerpretation. Amongst all the available maps, the GSWA geological lineament density map shows the lowest correlation with the hydrothermal mineral occurrences.

\section{Discussion}
\subsection{Evaluation of semi-automated computer vision framework}
In general, evaluation and validation of the different operations used in this study, against manual photointerpretation of the geological mapping, reveals an optimum dimension reduction technique, convolutional filter and edge detection parameter values. Each of these processes are discussed in turn.\\
Manual photointerpretation of the digital satellite data indicates that geological mapping by the GSWA captured most, but not all, of the geological lineaments in the Yinnetharra region of the Gascoyne Province (Fig. \ref{figure08}). Comparison of the extracted geological lineaments, using the proposed framework with manual photointerpretation and geological lineament map by the GSWA, reveals that applying MNF dimension reduction transformation is more reliable and results are closer to the field observations than the PCA and ICA (Fig. \ref{figure04}). The primary reason for the superior results from the MNF technique is high signal-to-noise ratio of the applied component \cite{Vassilas2002,Nielsen2011} and uniform behavior of each pixel's brightness value (positive peaks) facing the geological lineaments.\\
Visual (Fig. \ref{figure08}) and statistical (Fig. \ref{figure09}) comparisons show that the extracted geological lineaments using the proposed framework and applying a Laplacian filter on the MNF output component with the highest eigenvalue are well correlated with the GSWA geological lineament map. However, it is clear that the density map of the extracted geological lineaments using a directional filter is highly correlated with the hydrothermal mineral occurrences (Fig. \ref{figure10}). Moreover, the result of applying directional filter is similar to the map provided by the manual photointerpretation. The directional filter is able to enhance lineaments in desired directions which is an advantage in geological studies compared to the Laplacian filter. As geologists have prior knowledge of major structural trends, this approach is valid in regions where sufficient outcrop exists for field-based studies. Additionally, the images obtained by applying the directional filter significantly increases the efficiency of lineament extraction algorithms particularly through detecting small lineaments \cite{Lopes2016}. The broader spread of extracted lineament data using both filters and the manual photointerpretation compared to those extracted using the GSWA map is attributed to the short length of the extracted geological lineaments (Fig. \ref{figure09}) \cite{Adiri2017}.\\
The Canny edge detection and pixel connectivity-edge linking algorithms have been widely used in different fields \cite{Hashim2013,Ali2001,Liu2004,Elmahdy2016}. In this study, we increased the efficiency of these methods for extracting geological lineaments using the proposed framework. The initial steps of the proposed framework including dimension reduction techniques, adaptive and convolutional filters enhance the edges and increase the efficiency of edge detection and line extraction processes. Moreover, we present an optimum range of parameters applied in the lineament extraction operation for the Yinnetharra region (Table \ref{table01}). Optimum values of mentioned parameters may be different for each study area because of different spectral characteristics in each region, but our values provide a baseline for future studies. According to the field observations and geological evidence such as hydrothermal mineral occurrences, which are highly correlated with the structural features, our proposed framework is able to extract a high fraction of the geological lineaments (Fig. \ref{figure08}). A density map of the extracted geological lineaments using the proposed framework can be provided as the final product and can be used as an efficient evidential layer for prospecting hydrothermal mineral deposits.\\
Geological lineament maps provided by field observations are biased by the mapping geologist, who may have different views in dealing with the same structural features. The proposed framework is significantly less biased by the operator, but it is possible to modify the final map by applying new parameters to approach the desired geological lineament map. Ultimately, although it is always important to validate results whenever possible to geological maps, additional lineaments may be revealed through via a semi-automated computerized framework.

\subsection{Limitations of the proposed framework and future improvements}
The lineament extraction process applied in this study is limited to a specific curve length, while the geological lineaments show variable lengths on the Earth's surface. This semi-automated process extracts geological lineaments as multipart lines, which may be significantly in error with real structural discontinuities. Although structural discontinuities may naturally segmented by fault relays, longer, continuous faults are also likely to be present \cite{Vermilye1999}. In terranes where regolith or sedimentary cover is a problem, it is often difficult to differentiate between linked and single faults and, consequently, most geologists opt for producing single, longer faults as these require less inference \cite{Taylor2001,Fondriest2015}. Applied edge detection and line extraction methods in the proposed framework can be replaced with recently introduced methods based on image segmentation for other applications, to overcome above mentioned limitations \cite{Xiaoqi2016}.\\
The process of removing streams from the extracted lineaments using the proposed framework can also be improved by introducing new criteria through discriminating streams and geological lineaments. Here, an SRTM digital elevation model was used for mapping streams, but according to the Landsat 8 image (Fig. \ref{figure03}), some bifurcated streams failed to be identified using our semi-automated framework. High resolution images obtained using UAV-based photogrammetry can be considered as alternative input data to the proposed framework to overcome limited spatial resolution of the satellite data for efficient discrimination of streams and geological lineaments \cite{Vasuki2014}.\\
The proposed framework can be considered as a general methodology, which can be also used in other geoscientific fields such as hydrogeological and tectonic studies for mapping geological lineaments. We emphasize that maps provided using the proposed framework, which show the exposed geological lineaments on the Earth’s surface and deep-seated faults, can be further improved using other geophysical techniques, including magnetic, gravity and even radiometric data.

\section{Conclusion}
In this study, we present a computer vision-based framework for detecting geological lineaments using optical remote sensing data. The proposed framework involves diverse techniques for reducing dimensionality, removing noise and enhancing the lineaments in addition to edge detection and line extraction. The comparison of different dimension reduction techniques shows that the extracted geological lineaments using the output component of the MNF technique are best correlated with the available geological lineament map provided by the manual photointerpretation.\\
The Canny edge detector and pixel connectivity-edge linking algorithm are applied, respectively, to detect edges and to extract lines through providing a lineament map. The SRTM digital elevation model is applied to remove streams from the extracted lineaments to produce a geological lineament map. The extracted geological lineaments are compared to ground truth. Irrespective of the applied dimension reduction technique and convolutional filter, we observe a high fraction of the extracted geological lineaments are matched to the GSWA geological lineament map.\\
We also investigate the correlation between the density map of the extracted geological lineaments and known hydrothermal mineral occurrences in the study area. We conclude that the geological lineaments which are extracted using a directional filter are highly correlated with the hydrothermal mineral occurrences, even more than the map provided by the manual photointerpretation. We demonstrate that the output of the proposed framework can be applied to create an efficient evidential layer for prospectivity mapping of the hydrothermal mineral deposits.\\
In future work, the proposed framework can be used to map geological lineaments in other regions. The incorporation of various sources of data such as other types of geophysical maps (e.g., gravity, magnetics, and radiometrics) can be used to further enhance the framework. Moreover, statistical methods can be used to optimize parameters applied in lineament extraction algorithms. This could lead to better the convergence of the rose diagram obtained from the extracted geological lineaments and the field observations which are close to the real data.

\bibliographystyle{ieeetr}

\bibliography{References}

\end{document}